\documentclass[runningheads]{llncs}
\usepackage{amssymb}
\usepackage{amsmath}
\DeclareMathOperator{\minimize}{}
\usepackage{graphicx}
\begin{document}
\title{Explainable Causal Analysis of Mental Health on Social Media Data}
%
%
\author{Chandni Saxena\inst{1}
\and
Muskan Garg\inst{2}
\and
Gunjan Ansari \inst{3}
}
\institute{$^{1}$The Chinese University of Hong Kong, Hong Kong SAR\\
\email{csaxena@cse.cuhk.edu.hk}\\$^{2}$University of Florida, Gainesville, Florida, USA\\ \email{muskangarg@ufl.edu}\\ $^{3}$JSS Academy of Technical Education, Noida, India\\
\email{gunjanansari@jssaten.ac.in}
}
%
\authorrunning{C.~Saxena et al.}
\maketitle              
\begin{abstract}
With recent developments in \textit{Social Computing}, \textit{Natural Language Processing} and \textit{Clinical Psychology}, the social NLP research community addresses the challenge of automation in mental illness on social media. A recent extension to the problem of multi-class classification of mental health issues is to identify the cause behind the user's intention. However, multi-class causal categorization for mental health issues on social media has a major challenge of wrong prediction due to the overlapping problem of causal explanations. There are two possible mitigation techniques to solve this problem: (i) Inconsistency among causal explanations/ inappropriate human-annotated inferences in the dataset, (ii) in-depth analysis of arguments and stances in self-reported text using discourse analysis. In this research work, we hypothesise that if there exists the inconsistency among F1 scores of different classes, there must be inconsistency among corresponding causal explanations as well. In this task, we fine tune the classifiers and find explanations for multi-class causal categorization of mental illness on social media with LIME and Integrated Gradient (IG) methods. We test our methods with CAMS dataset and validate with annotated interpretations. A key contribution of this research work is to find the reason behind inconsistency in accuracy of multi-class causal categorization. The effectiveness of our methods is evident with the results obtained having category-wise average scores of $81.29 \%$  and $0.906$ using cosine similarity and word mover's distance, respectively. 

\keywords{ causal analysis \and explainability  \and mental health \and text categorization}
\end{abstract}
\section{Introduction}
People express their thoughts more conveniently on social media than during in-person (often analytical) sessions with experts. As per the National Institute of Mental Health report of 2020 \footnote{https://www.nami.org/mhstats}, 52.9 million adults in the USA suffer from mental illness. "The Health at a Glance Europe 2020" report\footnote{https://health.ec.europa.eu/system/files/2020-12/2020\_healthatglance\_rep\_en\_0.pdf} noted that the COVID-19 pandemic and the subsequent economic crisis caused a growing burden on the mental well-being of the citizens, with evidence of higher rates of stress, anxiety and depression. Previous studies support social media's powerful role in measuring the public's social well-being~\cite{robinson2019measuring}. To this end, we obtain Reddit social media posts demonstrating mental health issues for mental health analysis.

In this research work, we narrow down the problem of \textit{mental health analysis} to \textit{the identification of reasons behind users' intent in their social media posts}. The sequence to sequence (Seq2Seq) models are applied to solve the problem of causal categorization over CAMS dataset\footnote{https://github.com/drmuskangarg/CAMS}. The ground-truth of CAMS dataset contains two-fold annotations (i) \textit{causal category} and (ii) \textit{interpretations}. The textual segments of \textit{interpretation} support decision making for identifying causal categories. However, there exists a major challenge of responsibility and explainability for multi-class causal analysis while applying fine-tuned Seq2Seq models. In this context, we find explanations for inconsistency among resulting accuracy of different classes/ categories. Another key contribution is to find distance among \textit{inferences} and \textit{explanations} to obtain semantic similarity over distributional word representation: (i) \textit{cosine similarity} and (ii) \textit{word mover distance}. \\

\textbf{Definition 1: Inferences -} 
The inferences are set of interpreted textual segments by trained human-annotators which appears as ground-truth information in CAMS dataset.\\

\textbf{Definition 2: Explanations -}
The results obtained as the set of top-keywords using explainable AI approaches for multi-class causal categorization of Reddit posts is termed as explanations.\\

We further discuss a potential instance to define this problem of explainable causal analysis in this section. Consider a given sample $A$ where a user $U$ post $A$: ``\textit{Five years now and still no job. I am done with my life.}" The user $U$ is upset about his financial problems/ career due to \textit{unemployment}. We consider this text as the user-generated social media data which demonstrates mental health issues. The intent of a user is \textit{`to end life'} and a key challenge is to find the reason behind this intent. This cause-and-effect relationship aids the causal categorization. The category for sample $A$ is identified as `\textit{Jobs and careers}' because the reason is associated with unemployment. There are five causal categories in annotated CAMS dataset, namely, \textit{(i) bias or abuse, (ii) jobs and careers, (iii) medication, (iv) relationships, and (v) alienation}. 

In this research work, we use the CAMS dataset for explanations on multi-class causal categorization. We have made three major contributions in this work. First, we fine-tune deep learning models for multi-class causal categorization. Second, we obtain explainable text for causal categorization using \textit{Local Interpretable Model-Agnostic Explanations (LIME)} and \textit{IG}. Third, two semantic similarity measures: \textit{cosine similarity} and \textit{word mover distance} assist the validation of resulting explainable snippets with annotated inferences. Our experimental results explains the inconsistency among accuracy of different classes and validates the consistency of inferences made by model and human annotators, thereby defining the need of discourses and pragmatics for this problem of causal analysis. 

\section{Background}
Our task is defined as a domain-specific problem to find \textit{reasons behind the intent of a user on social media}. After extensive literature surveys, we observe minimal work on this problem. A domain-specific dataset is available for public use to examine the inferences (reasons) and causal categories (multi-class classification) task for mental health data as CAMS dataset~\cite{muskan2022lrec}. The existing solution of a task of causal analysis is given as the use of machine learning and neural models for multi-class categorization of causal categories. The resulting values of f-measure vary for different classes and raise a new research question: \textit{To what extent causal categorization is responsible?} We choose to resolve this problem by finding and validating the explainable texts.

To find the explanations for causal categorization, we explore existing explainable AI methods for natural language processing~\cite{madsen2021post}. Some well-established surveys and tutorials categorize explainable approaches into \textit{local vs global}, \textit{post hoc vs self explaining} and \textit{model agnostic vs model specific}~\cite{danilevsky2020survey}. We choose to observe local explanations with given input features for post-hoc interpretability methods which require less information. To this end, we identify two explainability approaches which are suitable for this study: (i) LIME and (ii) IG. 

LIME samples nearby observations and uses model estimates to fit the logistic regression~\cite{ribeiro2016should}. The parameters of logistic regression represent the importance measure and larger the parameters, greater effect will have on the output. The IG is an attempt to assign an attribution value to each input feature which measures the extent to which an input contributes to the final prediction~\cite{sundararajan2017axiomatic}. A recent study is carried out to set a benchmark over three representative NLP tasks (sentiment analysis, textual similarity and reading comprehension) for interpretability of both neural models and saliency methods~\cite{wang2022fine} thereby emphasizing the need of LIME and IG for downstream NLP tasks.  

The explainable methods give output in the form of important words/ text segments which serve as the most important input features. As we have available human annotated inferences for causal categorization in the form of text, we use these inferences as ground truth information (text-reference) and resulting explanations (RE) (text-observation). Thus, we use two semantic similarity measures to evaluate the performance of explainable methods for causal categorization- Cosine similarity and Word Mover's distance (WMD). Cosine similarity~\cite{salton1988term} calculates similarity between two words, sentences, paragraph, piece of text etc and evolves from the squared Euclidean distance measure which is used to measure how similar the documents are irrespective of their size. Word Mover's Distance (WMD) outperforms Bag-of-words and TF-IDF in terms of document classification error rates~\cite{kusner2015word}

\section{Framework}
In this section, we give a brief overview of the proposed framework for Figure~1 which represents the workflow for explainable causal analysis of mental health on social media data. We bifurcate our framework into three phases:
\begin{itemize}
    \item Causal Categorization: The use of neural models for causal categorization of Reddit posts depicting mental illness.
    \item Explanations: Finding explanations in the form of text-observations and obtaining top-keywords.
    \item Evaluations: Validate the resulting text-observations by comparing them with the human annotations available in the CAMS dataset. 
\end{itemize}

Consider a given set of self-reported short-text documents as $D$ where $D={d_1, d_2, ... , d_n}$. In \textit{Phase 1: causal categorization}, we segregate $D$ into training, validation and test set and give training set as an input and we fine-tune the multi-class classifier build model for our task. The model prediction are given as an input to \textit{Phase 2: Finding explanations} along with Reddit posts to obtain explanations. We further obtain these resulting explanations and human-annotated inferences present in the CAMS dataset for \textit{Phase 3: Evaluations} to test and validate the resulting explanations. Furthermore, we discuss three phases of our proposed framework in this section.


\begin{figure}
  \begin{center}
  
  \includegraphics[scale=0.9]{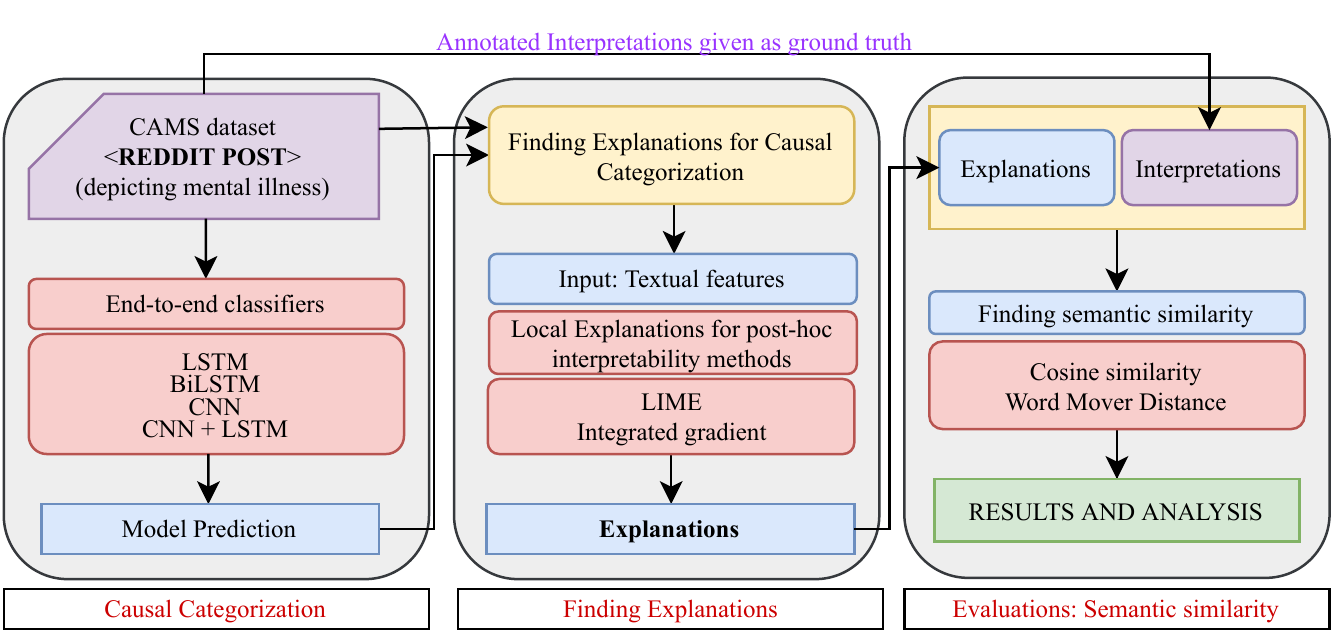}
   \label{fig:fig1}
     \caption{Overview of the proposed framework for explainable causal analysis of mental health on social media data. The framework is divided into three phases - Phase 1: Causal categorization, Phase 2: Finding explanations, Phase 3- Semantic similarity.}
  \end{center}
\end{figure}

\subsection{Phase 1: Causal categorization}
To solve the problem of causal categorization, we employ four learning based multi-class classifiers~\cite{chen2018deep}. We exploit following deep learning models and fine-tuned them for prediction:

\begin{itemize}
    \item \textbf{LSTM.} 
Long-Short Term Memory (LSTM:) is a popular advanced Recurrent neural network architecture for modeling sequential data which allows the information to persist and is trained by taking the sequence of the embedding feature vector.
    \item \textbf{BiLSTM:} A Bidirectional LSTM trains two hidden layers on the input sequence. The additional layer reverses the direction of information flow which means that the input sequence flows backward in an additional LSTM layer. 
    \item \textbf{CNN:} The CNN model efficiently extracts higher level features of the text using convolutional layers and max-pooling layers.
    \item \textbf{CNN-LSTM:} A Hybrid CNN-LSTM Model uses CNN layers for feature extraction on input text combined with LSTMs to support sequence prediction.
\end{itemize}

\subsection{Phase 2: Finding explanations}

We obtain local explanations by using following two post-hoc interpretability models:
\begin{itemize}
    \item \textbf{LIME:} It is a popular model-agnostic explainable method~\cite{ribeiro2016should} which provides local explanations for predictions of black-box models.LIME is also known as a post-hoc method. For a given model $\dot{F}$ and a given data sample $\alpha$, the method generates a fake dataset $ \alpha 1, \alpha 2, \alpha 3.. \alpha n$ and uses the black box model, $\dot{F}$ to obtain the target class or value for each sample. Subsequently, a white box model, $\bar{G}$ is trained with the generated data set along with the generated target labels. The aim is to train a white-box model for the original data sample and areas close to it even if the model does not perform as well globally. The closeness can be estimated using an appropriate similarity or distance metric. LIME then explains the original example using the white-box model and weights generated by it. The prediction accuracy of the white-box model, $\bar{G}$ gives an estimate of how close it mimics the black-box model, $\dot{F}$ and whether its explanations can be trusted. \\

    \item \textbf{Integrated Gradient:} The second method employed for explainability in this work is Integrated Gradients~\cite{sundararajan2017axiomatic}, a gradient-based explanation method. It is a model specific method that uses gradients (for example, using a deep neural network) to assess the importance of a feature on the model’s output. It employs the knowledge associated with the internal model for calculating the gradients of the model’s layers. It computes an attribution score corresponding to each feature by considering the integral of the gradients calculated along a straight path from a baseline instance $u{}'$ to the input instance $u$. 
\end{itemize}

\subsection{Phase 3: Evaluations with semantic similarity}
The \textit{human-annotated inferences} in CAMS dataset, which represents the causal explanation in the post, is validated by a senior clinical psychologist and it serves as a ground truth for our predicted explanations. There are two types of similarity measures for identifying document similarity (i) syntactic similarity and (ii) semantic similarity. We omit exact string matching algorithms due to varying number of words in each chunk and inconsistency among length of inferences and resulting keywords. The semantic similarity among texts validates the effectiveness of classifiers. We employ two most widely used semantic similarity measures: 
\begin{itemize}
    \item \textbf{Cosine similarity:} It is a widely used metric in information retrieval which models text as vector of terms~\cite{salton1988term}. The similarity of two input sentences (documents) can be derived by calculating cosine values of term vectors for the given input using the following equation. The similarity between two vectors of given input documents ($Doc_1 \, Doc_2$) can be defined as:
    \begin{equation}
    \label{eq1}
      Sim(Doc_1, Doc_2) = \frac{Doc_1 \cdot Doc_2}{|| Doc_1|| \,  ||Doc_2||} = \frac{\sum_{i=1}^{n}A_i \cdot B_i} {\sqrt{\sum_{i=1}^{n}A _i ^{2}} \sqrt{\sum_{i=1}^{n}B_i ^{2}}}
    \end{equation}
    where $A_i$ and $B_i$ represent the components of vectors $Doc_1$ and $Doc_2$, respectively.\\ 
    
    \item \textbf{Word Mover's Distance(WMD):} It is  a novel distance metric~\cite{kusner2015word} that is used to measure the dissimilarity between two text documents. The method is different from the conventional models that work on syntactic similarity rather than semantic similarity. The method employs word embedding like Glove and Word2Vec  to learn semantically meaningful representations of sentences. It computes distance between two documents A and B as the minimum cumulative distance that the embedded words of document A need to travel to reach the embedded words of document B.  
    
    WMD is computed using the cost-matrix having $x_i$ and $x_j$ be embedding of word i and j. The cost matrix $CM\in \mathbb{R}^{m}\times  \mathbb{R}^{m}$ is the distance of embeddings, such that $CM_{ij}= ||x_i - x_j ||^{2}$ as referred to in Eq.~\ref{eq2}.  The distance between two documents $Doc_1$ and $Doc_2$ is the optimum value of the following
    problem:
    \begin{equation}
    \label{eq2}
        \displaystyle{\minimize_{P \in \mathbb{R}^{m \times m}}  \, \sum_{ij}CM_{ij}\, P_{ij}}
    \end{equation}
    such that $P_{ij} \geq 0$
    Intuitively, $P_{ij}$ represents the amount of word $i$ that is transported to word $j$. WMD is defined as the minimum total distance to convert one document to another document.  
\end{itemize}

\section{Experiments and Evaluation}
This section covers the dataset description, experimental setup, results and performance evaluation of the proposed study.
\subsection{CAMS Dataset}
CAMS dataset consists of 5051 instances (1896 from SDCNL dataset and (ii) 3155 Reddit posts which are available
with subreddit r/depression using Python Reddit API
Wrapper (PRAW)\footnote{https://praw.readthedocs.io/en/stable/}) to categorize the \textit{direct causes} of mental disorders through mentions by users in their posts. Annotation is carried out manually by annotators who are proficient in the language. They work independently for each post and follow the given guidelines. Each annotator takes one hour to annotate about $15-25$ Reddit posts. The annotated files are verified by a clinical psychologist and a rehabilitation counselor. Furthermore, the validation of three annotated files is carried out by Fliess' Kappa inter-observer agreement study. The trained annotators have $61.28\%$ agreement for annotations of CAMS dataset. Despite the increased subjectivity of the task, the trained annotators \textit{substantially agree} with their judgements.

\subsection{Experimental Setup}
Considering a CAMS dataset, we divide it into training, validation and testing set consiting of $1699$, $117$ and $370$ instances, respectively. After preprocessing of the given documents $D$ (Reddit posts), we employ four deep learning methods to predict the causal category, namely, LSTM, BiLSTM, CNN and CNN-LSTM. At the initial layer of the neural network, we use GloVe, a distributional word embedding with dimension vectors of $100$. The GloVe embedding extracts semantics by using information available in neighbouring spaces. 
For experimental study, we consider a batch size of $128$, trained on $20$ epochs, with $265$ maximum length of tokens. We use Adam optimizer for all models with one or more dropout layers and optimal learning rate. We fine-tune CNN-LSTM model with a learning rate of $0.0005$ and set a learning rate of $1.46*10^{-3}$ for all other classifiers. 

\subsection{Experimental Results}
We perform experiments over the given dataset and obtain results as shown in Figure 2. To illustrate the effectiveness of our models, we give explanations for self-reported text of each causal-category. The given input is a \textit{self-reported text} of the CAMS dataset. The human-annotations are two-fold: (i) human-annotated interpretations (inferences) and (ii) causal category. We further perform explainable causal categorization to compare and contrast the \textit{inferences} with resulting top-keywords (\textit{explanations}). We observe minimal connection among words for Cause 0: No reasons followed by Cause 3: medications. However, the other causal categories seem to have high similarity among inferences and explanations.

\begin{figure}[t]
  \begin{center}
 
  \includegraphics[scale=0.99]{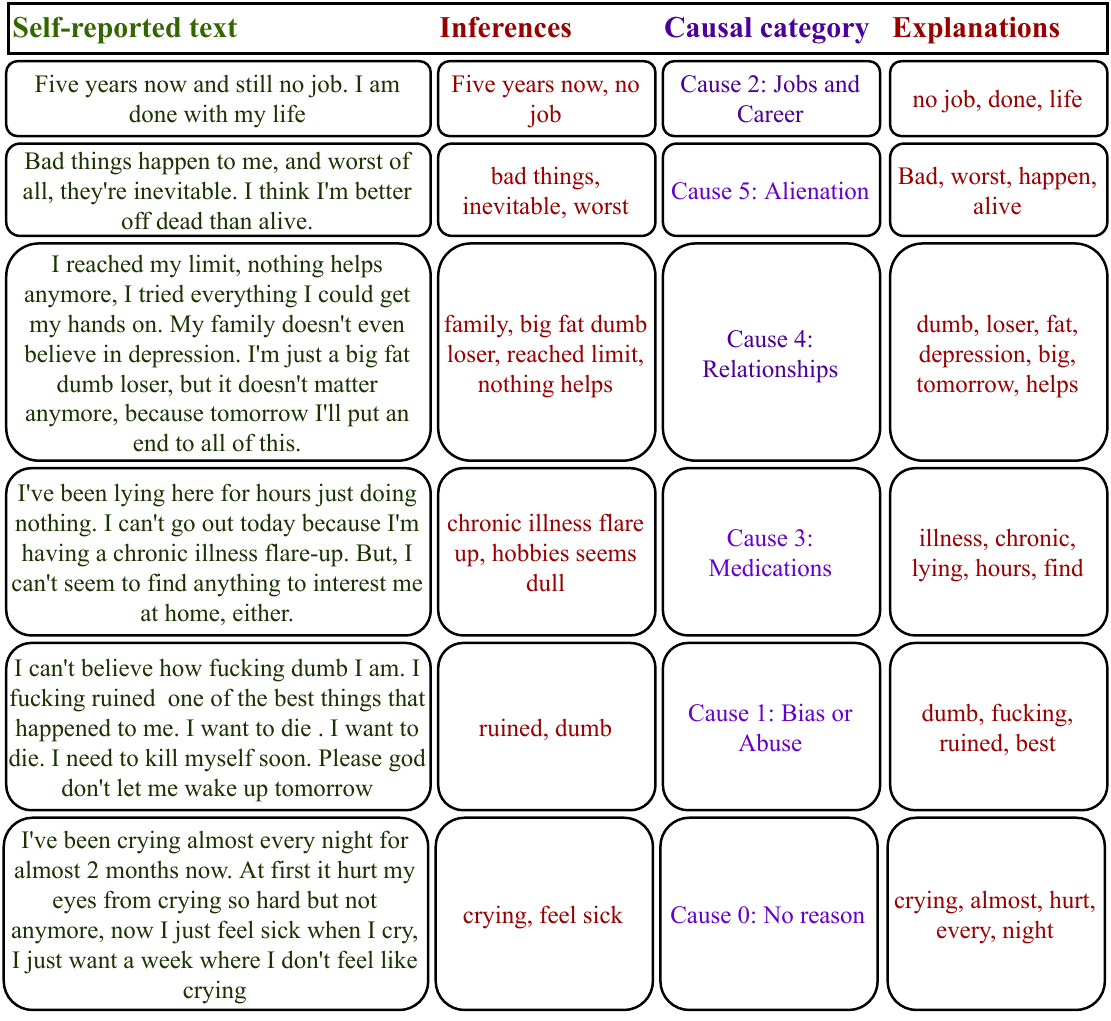}
   \label{fig2}
     \caption{Experimental results for explainable causal categorization for six different categories.}
  \end{center}
\end{figure}

\subsubsection{Error Analysis}
Other than the examples given in Figure 2, the medical terms mentioned in inferences and explanations may vary. For instance, prescriptions like \textit{propranalol}, name of diseases, heart problems, specific type of cancer and other antidepressants. This variation induces mismatch in semantic similarity among inferences and explanations for class 3.

\subsection{Performance Evaluation}
We use performance evaluation measures of the confusion matrix to evaluate the results of multi-class categorization. We analyze the results for each causal category and find overall accuracy of the model. Furthermore, we use two evaluation metrics of finding semantic similarity to evaluate explanations obtained by LIME and IG. 

\subsubsection{Causal categorization:}
We categorize the text into one of the six categories as mentioned in experimental results section and present the resulting values for multi-class classifiers in Table~\ref{tab1}. We observe the inconsistency in results for among different classes but consistency in variation among classes for different classifiers. To this end, we observe lowest F1 scores for causal category 1: \textit{Bias or Abuse}. The demonstration indicates errors among predictions for \textit{Alienation}/ \textit{Relationship}  as they overlap with \textit{Bias or Abuse}. The complex interactions illustrated the perceivable overlap between 
    \textit{Bias or Abuse} 
    and
    \textit{Relationship} in the following example:
    \begin{quote}
    \centering
       \textit{My friends are ignoring me and I am feeling bad about it. I have lost all my friends and don't want to live anymore.}
     \end{quote}
    The given example is associated with \textit{biasing} and \textit{friendship}, 
    in a case where someone feels ostracized by their friends.
    The emphasis on \textit{friends} tips the balance in favor of the class \textit{Relationship}. However, the major challenge is to train the model in such a way that it understands the inferences and then chooses the most emphasized \textit{causal category} using optimization techniques. 
    We view this challenge as an open research direction.

\begin{table}[tbp]
\begin{center}
\caption{Performance evaluation of multi-class classifiers for causal categorization of mental illness on social media data where F1:C0, F1: C1, F1:C2, F1:C3, F1:C4 and F1:C5 defines F1-score for 6 categories: cause 0: `No reason', cause 1: `Bias or abuse', cause 2: `jobs and careers', cause 3: `medication', cause 4: `relationships', and cause 5: `alienation', respectvely.}
\label{tab1}
\centering
\begin{tabular}{c|cccccc|c}
\hline
\textbf{Classifier} &\textbf{F1:C0} & \textbf{F1:C1} & \textbf{F1:C2} &\textbf{F1:C3} &\textbf{F1:C4} &\textbf{F1:C5} &\textbf{Accuracy}\\
\hline
LSTM  & 0.55  & 0.30  & 0.36  & 0.45   &   0.55 &  0.25 &  0.4514 \\
BiLSTM & 0.59 & 0.25  & 0.53  & 0.44   & 0.58  & 0.43 & 0.5054  \\
CNN  & 0.57 &  0.26 & 0.53 & 0.54 & 0.58 & 0.35  & 0.4919 \\
CNN-LSTM  & 0.57  & 0.17  & 0.38  & 0.46  &  0.48 & 0.52 & 0.4784\\
\hline
\end{tabular}
\end{center}
\end{table}

There are two possible mitigation techniques to solve this problem: (i) Inconsistency among causal explanations/ inappropriate human-annotated inferences in the dataset, (ii) in-depth analysis of arguments and instances in self-reported text using discourse analysis. In this research work, we hypothesise that if there exists the inconsistency among F1 scores of different classes, there exists an inconsistency among corresponding causal explanations as well. We find causal explanations and validate the results with human-annotated inferences. To this end, we choose to handle the first mitigation approach, thereby, enlisting new frontiers.  

\subsubsection{Explainability:}
In this section, we present evaluation of resulting top-keywords using LIME and IG methods. We use\textit{ word mover distance} and \textit{cosine similarity} over distributional word representations of both inferences and resulting keywords. As observed in Table~\ref{tab2}, the explanations of \textit{Class 0: 'No reason' }have maximum distance from human-annotated inferences for all methods. The reason is well-justified with the fact that Reddit posts having no reason behind intent of a user may or may not choose random words from the entire text. These random words does not describe any reason and thus, are the most far away from human-generated inferences. Low values for all other classes signifies the presence of patterns among explanations for other classes. We find \textit{class 2: jobs and careers}, and \textit{class 4: relationships} as the semantically most similar explanations achieved by deep learning methods.

\begin{table}[htbp]
\begin{center}

\caption{Values obtained for semantic similarity among resulting top-keywords and human-annotated inferences using \textit{Word Mover Distance}: More distance indicates less similarity among two different texts. }\label{tab2}
\centering
\begin{tabular}{p{7pc}|p{3pc}p{3pc}p{3pc}p{3pc}p{3pc}p{3pc}}
 
\hline
\textbf{Method used} &\textbf{Class0} & \textbf{Class1} & \textbf{Class2} &\textbf{Class3} &\textbf{Class4} &\textbf{Class5} \\

\hline
LSTM+LIME  & 1.029 & 0.854 & 0.857  & 0.896 & \textbf{0.838} & 0.889  \\
LSTM+IG & 1.097 & 0.890  & 0.870  & 0.926 & \textbf{0.867} & 0.906  \\
BiLSTM+LIME & 1.029  & 0.880 & 0.865 & 0.886 & \textbf{0.852} & 0.876 \\

BiLSTM+IG & 1.117  & 0.900 & 0.898   & 0.919 & \textbf{0.870} & 0.908 \\

CNN+LIME & 1.042  & 0.820 & 0.831   & \textbf{0.817} & 0.823 & 0.843 \\
CNN+IG & 1.123 & 0.907 & 0.882 & 0.912 & \textbf{0.880} & 0.913 \\
CNN-LSTM+LIME & 1.018  & 0.843 & \textbf{0.831} & 0.848 & 0.851  & 0.863 \\
CNN-LSTM +IG & 1.117 & 0.913 & \textbf{0.869} & 0.918 & 0.874 & 0.890 \\
\hline

\end{tabular}

\end{center}
\end{table}

We further analyse the results for cosine similarity as shown in Table~\ref{tab3}. We give input as a string, tokenize the text, use GloVe word embeddings to obtain word vectors, and find the mean of word vectors (obtained for each token). Experimental results demonstrate \textit{class 2: jobs and careers}, and \textit{class 4: relationships} as the most similar explanations to the human-annotated inferences. \textit{Class 3: Medication }, being associated with medical terms are expected to be semantically least similar as we would need domain-specific distributional word representation for evaluation in this category. Thus, class 3 and class 0 are illustrating low scores as compared to other classes.
\begin{table}[htbp]
\begin{center}

\caption{Values obtained for semantic similarity among resulting top-keywords and human-annotated inferences using \textit{Cosine Similarity}: The distance lies between 0 and 1}\label{tab3}
\centering
\begin{tabular}{p{7pc}|p{3pc}p{3pc}p{3pc}p{3pc}p{3pc}p{3pc}}

\hline
\textbf{Method used} &\textbf{Class0} & \textbf{Class1} & \textbf{Class2} &\textbf{Class3} &\textbf{Class4} &\textbf{Class5} \\

\hline
LSTM+LIME  & 0.787 & 0.825 & \textbf{0.889}  & 0.751 & 0.881 & 0.854  \\
LSTM+IG & 0.723 & 0.779  & \textbf{0.870}  & 0.701 & 0.869 & 0.813  \\
BiLSTM+LIME & 0.784  & 0.821 & \textbf{0.881} & 0.751 & 0.867 & 0.857 \\
BiLSTM+IG & 0.716  & 0.773 & \textbf{0.866} & 0.709 & 0.865 & 0.814 \\
CNN+LIME & 0.776  & 0.835 & \textbf{0.898}   & 0.822 & 0.894 & 0.861 \\
CNN+IG & 0.729 & 0.765 & \textbf{0.863} & 0.689 & \textbf{0.863} & 0.818 \\
CNN-LSTM+LIME & 0.781  & 0.831 & \textbf{0.878}   & 0.811 & 0.868 & 0.852 \\
CNN-LSTM+IG &0.728  & 0.789  & 0.851 &0.690 &\textbf{0.870} &0.815 \\
\hline

\end{tabular}

\end{center}
\end{table}

\subsection{Ethical Considerations}
NLP researchers are responsible for transparency about computational research with sensitive data accessed during model design and deployment. We understand the significance of ethical issues while dealing with a delicate subject of mental health analysis. We use the publicly available dataset and do not plan to disclose any sensitive information about the stakeholders (social media users) thereby preserving the privacy of a user~\cite{conway2014ethical}.

We use publicly available pre-trained base models for our demonstration to avoid any ethical conflicts. We assure that we adhere to all ethical guidelines to solve this task. Development of fair AI technologies in mental healthcare supports unbiased clinical decision-making~\cite{uban2021explainability}. Our research work is fair and there is no intentional bias as we consider explainable causal categories for mental health on CAMS dataset.

\section{Conclusion and Future Scope}
We find the explanations for causal categorization of mental health in social media posts by using LIME and IG methods, followed by performance evaluation by using human-annotated inferences in CAMS dataset. We conclude our work with three key takeaways: (i) less variations among resulting values of all classes for causal explanations as compare to F1 scores in causal categorization validates the human-annotated interpretations for causal categorization; (ii) the results for \textit{Class 0: No reason} and \textit{Class 3: Medication} are least explainable due to randomization and the need of domain-specific analysis, respectively; (iii) the performance evaluation of explanations obtained using explainable NLP is possible with  with semantic similarity methods if human-annotated interpretations are predefined.  

One of the path-breaking work is performed for causal explanation on social media which is obtained in the form of text~\cite{son2018causal}. The authors mentioned the complexity of this problem and made an attempt to resolve this issue by using discourses. However, the experiments were performed over a limited amount of Facebook data (\textit{often referred as Causal Explanation Analysis (CEA) dataset}) to classify the texts containing causal explanations and thereby extracting causal explanations. Furthermore, the causal explanation detection takes place on CEA dataset by capturing the salient semantics of discourses contained in their keywords with a bottom graph-based word-level salient network~\cite{zuo2020towards}. In this context, we choose to propose domain-specific discourse relation embeddings~\cite{son2021discourse} as a potential future research direction of causal analysis.

\bibliographystyle{splncs04}
\bibliography{mybibliography}
%




\end{document}